\documentclass[onecolumn]{IEEEtran}
\usepackage{cite}
\usepackage{graphicx} 
\usepackage{pgf,tikz}
\usetikzlibrary{arrows}
\usepackage{hyperref}
\usepackage[tight,footnotesize]{subfigure}

\usepackage[cmex10]{amsmath}
\usepackage{algorithmic}

\newtheorem{lemma}{Lemma} 
\newtheorem{theorem}{Theorem} 
\newtheorem{definition}{Definition} 
\newtheorem{example}{Example}

\newcommand{\opt}{\mathrm{opt}}
\newcommand{\non}{\mathrm{non}}
 
\begin{document}

\title{On the Easiest and Hardest Fitness Functions}

\author{Jun~He,   Tianshi~Chen and Xin~Yao, 
\thanks{This work was supported by the EPSRC under Grant Nos. EP/I009809/1 and EP/I010297/1. Xin Yao was supported by a Royal Society Wolfson Research Merit Award and also by the NSFC under Grant No. 61329302. Tianshi Chen was supported by the NSFC under Grant Nos. 61100163 and 61221062.}
\thanks{Jun He   is with Department of Computer Science, Aberystwyth University, Aberystwyth, SY23 3DB, UK.}%
\thanks{Tianshi Chen is with State Key Laboratory of Computer Architecture, Institute of Computing Technology, Chinese Academy of Sciences, Beijing 100190,   China.}%
\thanks{Xin Yao is with CERCIA, School of Computer Science, University of Birmingham, Birmingham B15 2TT, UK.}
}

\maketitle

\begin{abstract}
The hardness of fitness functions is an important research  topic in the field of evolutionary computation. In theory, the study can help understanding the ability of evolutionary algorithms. In practice, the study may provide a guideline to the design of benchmarks. The aim of this paper is to answer the following research questions: Given a fitness function class, which   functions  are the easiest with respect to an evolutionary algorithm?  Which are the hardest? How are these functions constructed?  The paper provides   theoretical answers to these questions. The easiest   and hardest fitness functions are constructed for  an elitist (1+1) evolutionary algorithm to maximise a class of fitness functions with the same optima.  It is demonstrated that the unimodal functions are  the easiest and deceptive functions are the hardest in terms of the time-based fitness landscape.  The paper also reveals that in a fitness function class, the easiest function  to one algorithm may become the hardest to another algorithm,  and vice versa.  
\end{abstract}

\section{Introduction} 
 Which fitness functions are easy for an evolutionary algorithm (EA)
and which are not? This is an important research  topic in the field of evolutionary computation. In theory, the study of the hardness of fitness functions can help understanding the ability of EAs. In practice, the study may provide a guideline to the design of benchmarks. 
  Answers to the above questions vary as the scope of fitness functions  changes from all possible functions to a single function.  

The first scenario is to  consider all possible fitness functions. In this case No Free Lunch theorems~\cite{wolpert1997no,wolpert2005coevolutionary} have answered the question. The theorems claim that the performance of any two EAs are equivalent in terms of average performance. 

The second scenario is to consider a class of fitness functions with the same features, such as unimodal  functions versus multi-modal functions, or deceptive functions versus non-deceptive functions. However a multi-modal function may be  easy   to solve \cite{horn1995genetic}. A unimodal function may be difficult for certain EAs but easy for others~\cite{rudolph1996how}. A non-deceptive function  may be difficult to an EA \cite{vose1995scalability}, and a deceptive function may be easy \cite{wilson1991gaeasy}.  Few features are available to distinguish whether a function class is easy or hard for an EA. 

The third scenario is to consider a single fitness function. A popular approach is to develop a statistic measure to predict the hardness of a fitness function,  such as
fitness-distance correlation~\cite{jones1995fitness},  fitness
variance~\cite{radcliffe1995fitness}, and epistasis
variance~\cite{davidor1991epistasis}. Unfortunately it is intractable to design a measure that can predict the hardness of a function efficiently \cite{naudts2000comparison,he2007note}.

Different from the above three scenarios, an alternative scenario is considered in  the current paper: 
given an EA and a class of  fitness functions with the same optima, which function is the hardest within the class? Which is the easiest? And how to construct them?   Here the easiest function is referred to a function on which the runtime of the EA is the shortest; and the hardest  is a function  on which the runtime of the EA is the longest. Both are compared with other functions in the same class.  These questions   have never rigorously been answered before.

Our research aims at understanding the hardest and easiest fitness functions within a  function class, and helping design benchmarks.   The set of benchmarks usually  include several typical fitness functions, for example, easy, hard and `averagely hard' functions in the class. An EA has the best performance on the easiest function, and the worst performance on the hardest function.  We will focus on these two extreme cases in this paper.

 The paper is organised as follows: Section~\ref{secWork} describes related work.  Section~\ref{secFunctions} defines the easiest and hardest fitness functions, and  establishes   criteria of determining whether a  function is the easiest or the hardest. Section~\ref{secConstruction} constructs    the easiest and hardest functions.  Section~\ref{secMutual} discusses the mutual transformation between the easiest and hardest functions. Section~\ref{sec9} concludes the paper.

\section{Related Work}
\label{secWork}
The hardness of fitness functions (or called problem difficulty) has been studied over two decades.  
Normally a fitness function is said to be easy to an EA if the  runtime is polynomial on the function or hard if the runtime is exponential. How to characterize which fitness functions are easy or hard   was thought to be a
major challenge~\cite{naudts2000comparison}. 

One approach is to link  features of a fitness landscape  to the hardness of fitness functions. Several features have been investigated, for example, isolation, deception and multi-modality, ruggedness and neutrality.  A fitness landscape with isolation   is hard for EAs, but other characteristics may not be
related too much to the hardness of fitness functions~\cite{naudts2000comparison}.     A fully non-deceptive function may be
difficult for an EA \cite{vose1995scalability} but   
some deceptive functions  can be solved easily by an EA \cite{wilson1991gaeasy}.  
Some multi-modal functions may be easy to solve~\cite{horn1995genetic}, but the unimodal function like the `long path' problem~\cite{horn1995genetic} could be difficult for certain
EAs~\cite{rudolph1996how}. Few  features are universally  useful to distinguish between hard and easy fitness functions.

 Another approach  is to predict the hardness of a fitness function  through  a statistic measure.  Many measures are proposed, for example,
fitness-distance correlation~\cite{jones1995fitness}, correlation length
and operator correlation~\cite{manderick1991genetic}, fitness
variance~\cite{radcliffe1995fitness}, and epistasis
variance~\cite{davidor1991epistasis}. Nevertheless,  to compute  the exact value of   such measures usually is
exponential in the problem size due to the fact that the search space is exponentially large \cite{jansen2001classifications,naudts2000comparison,he2007note}.  
Inherent flaws also exist in the common
hardness measures such as epistasis variance, fitness-distance
correlation and epistasis correlation \cite{reeves1999predictive}. 
 
An alternative  theoretical approach is based on  fitness levels.  Hard fitness functions  are classified into two types: `wide gap' problems  and `long path' problems~\cite{he2003analysis,he2003towards}. For the `wide gap' type, the EA is trapped
at a fitness level, because there is a wide gap between that fitness level and higher fitness levels. For the `long-path' type, the EA has to take a long path to reach an
optimum. The behavior of EAs on these two problems are different~\cite{chen2009new,chen2010choosing}.

The research in the current paper is totally different from previous work. The   hardest and easiest functions are compared with other fitness functions within the same function class. The hardest function are not relevant to exponential runtime and the easiest fitness functions  are not relevant to polynomial runtime.    For some function class,   an EA  only needs  polynomial time on the hardest function. For some other function class, an EA may take exponential time on the easiest function.

Our study is  also different from No Free Lunch theorems \cite{wolpert1997no,wolpert2005coevolutionary}, which  state that any two EAs are equivalent when their performance is averaged across all possible fitness functions. We don't intend to investigate the easiest and hardest functions among all possible fitness functions, instead only within a class of fitness functions with the same optima.

\section{Easiest and Hardest Fitness Functions}
In this section we  define   the easiest   and    hardest fitness functions in a  function class  and establish  the criteria to determine whether a function is the easiest or hardest. 
\label{secFunctions}
\subsection{Definition of Easiest and Hardest Fitness Functions}
Consider the  problem of maximizing a class of fitness functions with the same optima. An instance of the problem is to maximize a fitness function $f(x)$:
\begin{equation}\label{equProblem}
  \max \{ f(x); x \in  S \},  
\end{equation}
where   $S$ is a finite set.   The optimal set is  denoted by $S_{\mathrm{opt}} $ and the   non-optimal set  by $S_{\mathrm{non}}$.  Without loss of generality, the function   $f(x)$ takes  $L+1$ finite values $f_0 > f_1 > \cdots > f_L$ (called \emph{fitness levels}). Corresponding to fitness levels, the set $S$ is decomposed into $L+1$ subsets: 
$$
S_l: =\{x \mid f(x) =f_l \}, l=0,1, \cdots, L.
$$

For simplicity of analysis, we only investigate  strictly elitist (1+1) EAs. Using  \emph{strictly elitist selection}, the  parent  is replaced by the  child   only   when the child   is fitter. Therefore the best found solution is always preserved.
In the EAs,  mutation is independent of the fitness function.  Both  mutation and selection operators are time invariant  (i.e., static).
 The  procedure of such  an elitist (1+1) EA is described as follows.   

\begin{algorithmic}[1]
\STATE \textbf{input}: fitness function $f(x)$;
\STATE generate a solution at random and denote it by $ \phi_0$;  
\STATE generation counter $ t\leftarrow 0$; 
\WHILE{ the maximum value  of $f(x)$ is not found }
\STATE child $\phi_{t.m}\leftarrow$ is mutated from parent $\phi_t$; 
\IF {$f(\phi_{t.m})> f(\phi_t)$ }
\STATE   next generation parent $\phi_{t+1} \leftarrow \phi_{t.m}$;
\ELSE
\STATE    next generation parent $\phi_{t+1} \leftarrow \phi_t$;
\ENDIF  
\STATE $t\leftarrow t+1$;
\ENDWHILE \STATE \textbf{output}:  the maximal value of $f(x)$.
\end{algorithmic}

Let $G(x)$ denote the expected number of generations  for an EA to find an  optimal solution for the first time when starting at $x$ (called \emph{expected hitting  time}). In (1+1) EAs,  $G(x)$ also represents   the expected number of fitness evaluations (called  \emph{expected runtime}).  In this paper, we  restrict our discussion to those EAs whose expected runtime is finite (convergent).

\begin{definition}
Given an EA for maximising a class of fitness functions with the same optima, a function $f(x)$ in the class is said to be the \emph{easiest} to the   EA if starting from any initial point,
the  runtime   of the EA for maximising  $f(x)$
is no  more than
the  runtime for maximising any fitness function  $g(x)$ in the class when   starting from the same initial point.
A   function $f(x)$ in the class is said to be the \emph{hardiest} to the  EA if starting from any initial point,
the  runtime  of the EA for maximising  $f(x)$
is no less than
the  runtime  for maximising any fitness function  $g(x)$ in the class when   starting from the same initial point.
\end{definition}

The definition of the easiest and hardest functions is based on a point-by-point comparison of the runtime of the EA on two fitness functions. It is irrelevant to polynomial or exponential runtime. The easiest  and hardest functions are not unique. This will be demonstrated in Subsection~\ref{secCase2}.

 \subsection{Criterion for Determining Easiest  Function}
 \label{secCriteriaEasiest} 
Before we establish the criterion, we apply drift analysis   to the random sequence $\{\phi_t, t=0, 1,\cdots\}$ and draw several preliminary results. Notice that each generation of the (1+1) EA consists of two steps: mutation and selection,   
\begin{align*}
\phi_t \overset{\text{mutation}}\longrightarrow \phi_{t.m} \mbox{ with } \phi_{t}  \overset{\text{selection}}\longrightarrow \phi_{t+1}.
\end{align*}

 The mutation operator is  a transition from   $\phi_t$ to  $\phi_{t.m}$, whose transition probabilities are represented by 
\begin{align}
P^{[m]}(x,y):=P(\phi_{t.m}=y \mid \phi_t =x), x, y \in S.
\end{align} 
Here $\phi$ is a random variable and $x$ its value. 

The selection operator is another transition from    $\phi_t$  and $\phi_{t.m}$ to  $\phi_{t+1}$, whose transition probabilities are represented by 
\begin{align*}
P^{[s]}(x,y; z):=P(\phi_{t+1}=z \mid \phi_t =x, \phi_{t.m}=y),  x, y,z \in S.
\end{align*}  

The $t$th generation is   a transition  from   $\phi_t$ to $\phi_{t+1}$, whose transition probabilities are represented by 
\begin{align}
P(x,z):=P(\phi_{t+1}=z \mid \phi_t =x).
\end{align}

In drift analysis, a function $d(x)$ is called a \emph{drift function} if it   is non-negative at any point and  equals to 0 at any optimum. Given a drift function $d(x)$, \emph{drift}  
  represents the progress rate  of  moving towards the optima per generation. Drift   at   point $x$  is  defined by
$$
\Delta(x): = \sum_{y \in S}   P(x,y)  ( d (x) -d (y)).
$$
   
Define positive drift $\Delta^+(x)$ and negative drift $\Delta^-(x)$ as follows 
 \begin{align*}
\Delta^+(x)    =& \sum_{y: d (x) > d (y)}  P ( x, y) ( d (x) -d (y)),  \\
 \Delta^- (x)    =& \sum_{y: d (x)< d (y)}  P ( x, y) ( d (x) -d (y)).
 \end{align*}
Then the  drift $
\Delta(x) = \Delta^+(x) +\Delta^-(x).
$

Using drift analysis~\cite{he2004study}, we obtain the following    preliminary  results.

\begin{lemma} \cite[Lemma 1]{he2004study}
\label{lemTimeUpperBound}
 If the drift satisfies that $\Delta  (x) \ge 1$ for any non-optimal point $x$, then    the expected runtime satisfies that  $G(x) \le d(x) $ for any   point $x$.
\end{lemma} 

\begin{lemma}  \cite[Lemma 2]{he2004study}
\label{lemTimeLowerBound} 
 If the drift satisfies that $\Delta  (x) \le 1$ for any non-optimal point $x$,  then    the expected runtime satisfies that  $G(x) \ge d(x) $ for any   point $x$.
\end{lemma}

\begin{lemma} \cite[Lemma 3]{he2004study}
\label{lemDriftOne}
 Let the drift function $d(x)=G(x)$, then the drift satisfies
$\Delta(x)=1$ for any non-optimal point $x$.
\end{lemma}

Furthermore, the runtime of an elitist (1+1) EA can be explicitly expressed in transition probabilities.
 
\begin{lemma} \cite[Theorem 4]{he2003towards} \label{lemExactTime}
For any  elitist (1+1) EA, its expected runtime is given by 
\begin{equation*}
G(x)=\left\{
\begin{array}{lll}
0,  &  x=S_0. \\
\frac{1+\sum^{l-1}_{k=0}\sum_{y \in S_k} P(x,y)G(y) }{ \sum^{l-1}_{k=0} \sum_{y \in S_k} P(x,y)}, 
 &x \in S_l, l >0.
\end{array}
\right.
\end{equation*}

\end{lemma}

Using the above lemmas, we   establish a criterion of determining whether a fitness function is the easiest  to an elitist   (1+1) EA.

\begin{theorem}\label{theCriterionEasy}
Given an elitist (1+1) EA, and  a class of fitness functions with the same optima, let $G_f(x)$ denote the runtime of the (1+1) EA for maximising $f(x)$. If  the following \emph{monotonically decreasing condition}  holds:
\begin{itemize}
\item for any two  points  $x$ and $y$ such that $G_f(x) < G_f(y)$, it has $f(x) > f(y)$,
\end{itemize} 
  then  $f(x)$ is the easiest in the fitness function class.
\end{theorem}

\begin{IEEEproof} 
Let $g(x)$ be a fitness function in the function class. $\{ \phi_t, t=1, 2, \cdots \}$ denotes the sequence for maximising $f(x)$, and $\{\psi_t, t=1,2, \cdots \}$ the sequence for maximising  $g(x)$.  Let $G_g(x)$ denote the runtime of the (1+1) EA for maximising $g(x)$.

Since our objective is to show the expected runtime on $f(x)$ is no more than the runtime on any other function, we   take the runtime on $f(x)$ as the drift function: $
d(x)= G_f (x)
$. This plays a crucial role in our analysis.

For the sequence $\{\phi_t \}$, denote the  drift at  point $x$  by $\Delta_{\phi}(x)$. For the sequence $\{\psi_t \}$, denote the  drift at  point $x$  by $\Delta_{\psi}(x)$.
The subscripts $\phi$ and $\psi$ are used to distinguish between the two sequences $\{\phi_t\}$ and $\{ \psi_t \}$.

Notice that $d(x) =G_f(x)$, then we apply Lemma~\ref{lemDriftOne} and get that for any non-optimal point $x$, drift
\begin{align}
\label{equDriftEq1}
\Delta_{\phi}(x)  =1.
\end{align}

The rest of proof is based on the idea: first, we prove the drift $\Delta_{\psi}(x) \le 1$ for the sequence  $\{\psi_t \}$, and then   draw the derived conclusion
using Lemma~\ref{lemTimeLowerBound}.

(1)  First we compare the negative drift of the two sequences. In the case of negative drift, we consider two points $x$ and $y$ such that $d(x) < d(y)$ (i.e., $G_f(x) < G_f(y)$).
 According to the monotonically decreasing condition,   $f(x) > f(y)$.

For the sequence $\{ \phi_t\}$, $y$ is never accepted due to elitist section, which leads to
$
P_{\phi}(x,y)=0.
$
Thus for the  sequence $\{ \phi_t\}$, there is no negative drift.
$$
\Delta^-_{\phi}  (x)=    0.
$$

For the sequence $\{ \psi_t\}$, there exist  two cases:  (i)  $g(x) < g(y)$;  (ii)  $g(x) \ge g(y)$. In the case  of  $g(x) < g(y)$,   $y$ will  be accepted, which implies
$
P_{\psi}(x,y)\ge 0.
$
Thus  there exists negative drift for the sequence $\{ \psi_t \}$.
$$
\Delta^-_{\psi}  (x) \le  0.
$$

Comparing the negative drift of these two sequences, we get 
\begin{align}
\label{equDriftNeg1}
\Delta^-_{\psi}  (x) \le \Delta^-_{\phi}  (x).
\end{align}

(2)  Secondly we compare the positive drift  of the two sequences. In the case of positive drift, we consider two points $x$ and $y$ such that $d(x) > d(y)$.
If $y$ is not an optimum, then according to the monotonically decreasing condition,    $f(x) < f(y)$. If $y$ is an  optimum, then naturally  $f(x) < f(y)$.

For the sequence $\{\phi_t \}$, if such a $y$ has been  mutated from $x$, then $y$ is always accepted due to elitist selection. Thus  
\begin{align*}
P_{\phi}(x,y)=P^{[m]}(x,y).
\end{align*}

For the sequence $\{\psi_t \}$,  there exist  two cases:  (i)  $g(x) < g(y)$;  (ii) or  $g(x) \ge g(y)$. 
In the case of $g(x) < g(y)$,   according to elitist section, $y$ is always accepted. Thus  
\begin{align*}
P_{\psi}(x,y)=P^{[m]}(x,y).
\end{align*} 

In the case of $g(x) \ge g(y)$,    according to elitist section, $y$ will not be    accepted.  
The transition probability  
 $ P_{\psi}( x, y)=0.$

Then we get that
$ P_{\phi}( x, y)  \ge P_{\psi}( x, y) .
$
Hence 
  \begin{align*}
 & \sum_{y: d(x) > d (y)}  P_{\psi}( x, y) ( d (x) -d(y)) \\
  \le & \sum_{y:d (x) >d (y)}  P_{\phi}( x, y) ( d (x) -d(y)).
 \end{align*}
So  the positive  drift  of  the two sequences satisfies
  \begin{align}
  \label{equDriftPos1}
  \Delta^+_{\psi}(x) \le \Delta^+_{\phi}   (x) .
  \end{align}

Merging   (\ref{equDriftNeg1}) and (\ref{equDriftPos1}) and   using (\ref{equDriftEq1}), we know that the total drift of the two sequences satisfies
$$
\Delta_{\psi} (x) \le \Delta_{\phi}(x)=1.
$$

Applying Lemma~\ref{lemTimeLowerBound}, we see  the expected runtime on $g(x)$ satisfies
$$
G_g(x) \ge d(x)=G_f(x) ,
$$
then we finish the proof.
\end{IEEEproof}

Now we give an intuitive explanation of the above theorem. The   monotonically decreasing condition means the function is unimodal in terms of the time-based fitness landscape and  Theorem~\ref{theCriterionEasy} asserts that a unimodal function is always the easiest. In the following we explain this in detail.

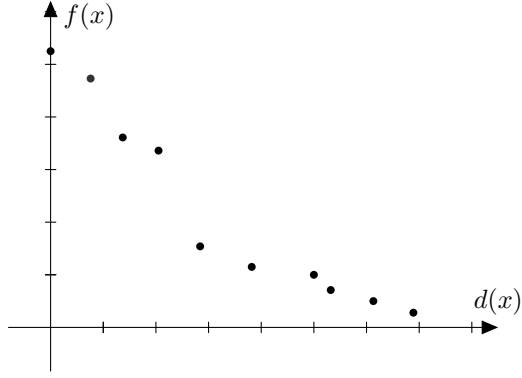
\begin{figure}[ht]
\centering
\definecolor{tttttt}{rgb}{0.2,0.2,0.2}
\begin{tikzpicture}[line cap=round,line join=round,>=triangle 45,x=0.7cm,y=0.7cm]
\draw[->,color=black] (-0.8,0) -- (8.5,0);
\foreach \x in {,1,2,3,4,5,6,7,8}
\draw[shift={(\x,0)},color=black] (0pt,2pt) -- (0pt,-2pt);
\draw[color=black] (7.86,0.06) node [anchor=south west] { $d(x)$};
\draw[->,color=black] (0,-0.83) -- (0,6.21);
\foreach \y in {,1,2,3,4,5,6}
\draw[shift={(0,\y)},color=black] (2pt,0pt) -- (-2pt,0pt);
\draw[color=black] (0.07,5.92) node [anchor=west] { $f(x)$};
\clip(-0.8,-0.83) rectangle (8.5,6.21);
\fill [color=black] (0,5.25) circle (1.5pt);
\fill [color=black] (2.84,1.54) circle (1.5pt);
\fill [color=black] (5,1) circle (1.5pt);
\fill [color=black] (6.89,0.28) circle (1.5pt);
\fill [color=tttttt] (0.76,4.73) circle (1.5pt);
\fill [color=black] (2.05,3.36) circle (1.5pt);
\fill [color=black] (1.37,3.61) circle (1.5pt);
\fill [color=black] (5.32,0.71) circle (1.5pt);
\fill [color=black] (3.82,1.15) circle (1.5pt);
\fill [color=black] (6.13,0.5) circle (1.5pt);
\end{tikzpicture}
\caption{A unimodal time-based fitness landscape. The $x$ axis is   the runtime: $d(x)=G(x)$. The $y$ axis is the fitness function. The origin represents the optimum.}
\label{figUniModal}
 \end{figure}

In a \emph{time-based fitness landscape}, runtime $G(x)$ is regarded as the distance $d(x)$ between a point $x$ and the optimum. It is completely different from a neighbourhood-based distance such as the Hamming distance. Time is seldom used as a   distance measure in evolutionary computation but popular in our real life.   Taking runtime as the distance, we visualise the   monotonically decreasing condition  
\begin{itemize}
\item for any two  points  $x$ and $y$ such that $d(x)< d(y)$, it has $f(x) > f(y)$,
\end{itemize} using a  time-based fitness landscape (see \figurename~\ref{figUniModal}), where the $x$ axis is the runtime and the $y$ axis is the fitness, and the origin   represents the set of optima with $d(x)=0$.  
 
 The landscape is  \emph{unimodal}:  the function $f(x)$ has exactly one   optimum. In contrast, any unimodal function defined in the 2-D time-based fitness landscape will satisfy the monotonically decreasing condition.  The unimodal property implies that no negative drift exists in an elitist EA. Thus the EA always moves towards the optimum. This makes the unimodal time-based fitness landscapes the easiest to the EA.

The theorem only states that a unimodal time-based fitness landscape is the easiest. Nevertheless this assertion could not be established if using a   neighbourhood-based distance such as the Hamming distance. A unimodal function in the context of a neighbourhood-based fitness landscape is not always the easiest.

 \subsection{Criterion of Determining Hardest  Function}
 \label{secCriteriaHardest}

In a similar way, we establish a criterion of determining whether a fitness function is the hardest  to an elitist  (1+1) EA. It is similar to Theorem~\ref{theCriterionEasy}. The    monotonically decreasing condition  is replaced by the monotonically increasing condition.

\begin{theorem}\label{theCriteronHard}
Given an elitist (1+1) EA, and  a class of fitness functions with the same optima, let $G_f(x)$  denote the expected runtime for maximising $f(x)$. If  the following \emph{monotonically increasing condition}  holds:
\begin{itemize}
\item for any two non-optimal points  $x$ and $y$ such that $G_f(x)< G_f(y)$, it has $f(x) < f(y)$,
\end{itemize} 
  then    $f(x)$ is the hardest in the class. 
\end{theorem}

\begin{IEEEproof} 
The proof is similar to that of Theorem~\ref{theCriterionEasy} but with several  changes.  

 Let $g(x)$ be a fitness function in the function class. $\{ \phi_t, t=1, 2, \cdots \}$ denotes the  sequence  for maximising $f(x)$, and $\{\psi_t, t=1,2, \cdots \}$ the sequence for maximising  $g(x)$.   $G_g(x)$ denotes the runtime of the (1+1) EA for maximising $g(x)$.
We  take the runtime on $f(x)$ as the drift function:
$
d(x)= G_f (x).
$

For the sequence $\{\phi_t \}$, notice that $d(x) =G_f(x)$, then we apply Lemma~\ref{lemDriftOne} and get for any non-optimal point $x$
\begin{align}
\label{equDriftEq2}
\Delta_{\phi}(x)  =1.
\end{align}

(1) First we compare the negative drift  of the two sequences. We consider two non-optimal points $x$ and $y$ such that $d(x) < d(y)$ (i.e., $G_f(x) < G_f(y)$). According to the monotonically increasing condition,  $f(x) < f(y)$.

For the sequence $\{\phi_t \}$, if such a $y$ has been  mutated from $x$, then $y$ is always accepted due to elitist selection. Thus  
\begin{align*}
P_{\phi}(x,y)=P^{[m]}(x,y).
\end{align*}

For the sequence $\{\psi_t \}$,  there exist  two cases:  (i)  $g(x) < g(y)$;  (ii) or   $g(x) \ge g(y)$. 
In the case of $g(x) < g(y)$,   if such a $y$ has been  mutated from $x$, then $y$ is always accepted due to elitist selection. Thus  
\begin{align*}
P_{\psi}(x,y)=P^{[m]}(x,y).
\end{align*} 

In the case of $g(x) \ge g(y)$,    according to elitist section, $y$ will not be    accepted.  
The probability $P_{\psi}( x, y)$ equals to
 $$ P_{\psi}( x, y)=0.$$

Then we get that
$
 P_{\phi}( x, y)  \ge P_{\psi}( x, y) .
$
Hence 
  \begin{align}
 \nonumber & \sum_{y: d(x) <d (y)}  P_{\psi}( x, y) ( d (x) -d(y)) \\
  \ge & \sum_{y:d (x) <d (y)}  P_{\phi}( x, y) ( d (x) -d(y)).
 \end{align}
Equivalently  the negative  drift of the two sequences satisfies
\begin{align}
\label{equDriftNeg2}
\Delta^-_{\psi}(x) \ge \Delta^-_{\phi}   (x) . 
\end{align}

(2)  Secondly we compare the positive drift  of the two sequences. We consider two points $x$ and $y$ such that $d(x) > d(y)$, where $y$ could be either an optimum or not.

First consider $y$ an optimum. For the sequence $\{\phi_t \}$, if such a $y$ has been  mutated from $x$, then $y$ is always accepted due to elitist selection. Thus  
\begin{align*}
P_{\phi}(x,y)=P^{[m]}(x,y).
\end{align*}

Similarly for the sequence $\{\psi_t \}$,  $y$ is always accepted due to elitist selection. Thus 
\begin{align*}
P_{\psi}(x,y)=P^{[m]}(x,y).
\end{align*}
Then we get 
\begin{align}
\label{equProOpt} 
P_{\psi}(x,y)=P_{\phi}(x,y).
\end{align}

Then consider $y$ not an optimum. According to the monotonically increasing condition,    $f(x) > f(y)$ if $ y$ is not an optimum. 
 
 For the sequence $\{\phi_t\}$, $y$ is never accepted due to elitist section, which leads to
\begin{align*}
P_{\phi}(x,y)=0.
\end{align*}

For the sequence $\{\psi_t\}$, even if $f(x) > f(y)$, it is still possible that  $g(x) < g(y)$. So $y$ may be accepted. This means 
\begin{align*}
P_{\psi}(x,y)\ge 0.
\end{align*}
Thus  we have
\begin{align}
\label{equProNon}
  P_{\psi}(x,y) \ge P_{\phi}(x,y).
\end{align}

Combining (\ref{equProOpt}) and (\ref{equProNon}), we have  for any $y$,
$$
 P_{\psi}(x,y) \ge P_{\phi}(x,y).
$$ 

Then 
\begin{align}
&\nonumber   \sum_{y:   d(x) > d(y)}  P_{\psi}( x, y) ( d (x) -d(y)) \\&\ge   \sum_{y : d(x) > d(y)}  P_{\phi}( x, y) ( d (x) -d(y)).
\end{align}
 Equivalently the positive drift of the two sequences satisfies
\begin{align}
\label{equDriftPos2}
\Delta^+_{\psi}  (x) \ge \Delta^+_{\phi}  (x).
\end{align}

Merging  (\ref{equDriftNeg2}) and  (\ref{equDriftPos2}) and using   (\ref{equDriftEq2}), we draw that the total drift of the two sequences satisfies
$$
\Delta_{\psi} (x) \ge \Delta_{\phi}(x)=1.
$$

It follows from Lemma~\ref{lemTimeUpperBound} that for any non-optimal point $x$
$$
G_g(x) \le d(x)=G_f(x) ,
$$
then we finish the proof.
\end{IEEEproof}

An intuitive explanation of the above theorem is that  the   monotonically increasing condition means the function is deceptive and  Theorem~\ref{theCriteronHard} states a deceptive function is always the hardest.  Let's demonstrate this using the time-based fitness landscape. Still taking runtime $G(x)$ as the distance $d(x)$ to the optima, we visualise the    monotonically increasing condition  
\begin{itemize}
\item for any two non-optimal points  $x$ and $y$ such that $d(x)< d(y)$, it has $f(x) < f(y)$,
\end{itemize}  using a time-based fitness landscape (see \figurename~\ref{figDeceptive}). The landscape is  \emph{deceptive}:  the closer a point is to the origin, the lower its fitness is.  The deceptive time-based fitness landscape is the hardest. 

\begin{figure}[ht]
\centering
\begin{tikzpicture}[line cap=round,line join=round,>=triangle 45,x=0.7cm,y=0.7cm]
\draw[->,color=black] (-0.8,0) -- (8.5,0);
\foreach \x in {,1,2,3,4,5,6,7,8}
\draw[shift={(\x,0)},color=black] (0pt,2pt) -- (0pt,-2pt);
\draw[color=black] (7.86,0.06) node [anchor=south west] { $d(x)$};
\draw[->,color=black] (0,-0.83) -- (0,6.21);
\foreach \y in {,1,2,3,4,5,6}
\draw[shift={(0,\y)},color=black] (2pt,0pt) -- (-2pt,0pt);
\draw[color=black] (0.07,5.92) node [anchor=west] { $f(x)$};
\clip(-0.8,-0.83) rectangle (8.5,6.21);
\fill [color=black] (0,5.33) circle (1.5pt);
\fill [color=black] (4.83,2.99) circle (1.5pt);
\fill [color=black] (5.37,3.9) circle (1.5pt);
\fill [color=black] (5.65,5.04) circle (1.5pt);
\fill [color=black] (3.79,1.47) circle (1.5pt);
\fill [color=black] (4.54,2.44) circle (1.5pt);
\fill [color=black] (4.11,1.84) circle (1.5pt);
\fill [color=black] (5.5,4.26) circle (1.5pt);
\fill [color=black] (5.09,3.45) circle (1.5pt);
\fill [color=black] (5.59,4.76) circle (1.5pt);
\end{tikzpicture}
  \caption{A deceptive time-based fitness landscape.   The $x$ axis is    the runtime: $d(x)=G(x)$. The $y$ axis is the fitness function.  Th origin represents the optimum.}
\label{figDeceptive}
\end{figure}
 
When  using a  neighbourhood-based distance, it is impossible to establish a similar result under a similar condition. A deceptive function in the context of the neighbourhood-based fitness landscape is not always the hardest. 

Note: the above unimodal and deceptive time-based fitness landscapes are  different from the easy and hard fitness landscapes described in \cite{he2004understand}, which are classified by polynomial or exponential hitting time.

\subsection{Case Study:  0-1 Knapsack Problem}
We give two simple examples to show the application of the above theorems. The examples come from  the 0-1 knapsack problem. We will not consider all instances of the 0-1 knapsack problem. Instead we   focus on an instance class.

\begin{example}
Consider an  instance class of the 0-1 knapsack problem described as follows:
\begin{equation}
\begin{array}{lll}
 \mbox{maximize } & f(x)=\sum^n_{i=1} v_i x_i, \nonumber\\
 \mbox{subject to }&\sum^n_{i=1} w_i x_i \le C, \nonumber
\end{array}
\end{equation}
where $v_i>0$ is the value of  item $i$,  $w_i>0$  its weight, and $C$ the knapsack capacity. The value of items satisfies $v_1 > v_2 +\cdots + v_n$, and the weight of items satisfies $w_1 > w_2+\cdots +w_n$, the knapsack capacity  $C =w_1$.  A solution  is represented by a binary string  $x=(x_1  \cdots x_n)$.  The unique optimum is $(1  0 \cdots 0)$, denoted by $x^*$.

An elitist (1+1) EA using bitwise mutation is applied to  the problem. 
\begin{itemize}
\item \textbf{EA($\frac{1}{n}$).} Flip each bit independently with flipping probability $\frac{1}{n}$.  
\end{itemize}
For simplicity of analysis,  we adopt the simplest approach to handle the constraint: reject any infeasible solution during selection. 

Let's investigate a special instance in the class: $v_1=n$ and $v_2= \cdots = v_n=1$;  $w_1=n$ and $w_2 =\cdots =w_n=1$.    
Notice that the global optimum is $x^*:=(1 0  \cdots 0)$ and the local optimum is $(0 1  \cdots  1)$. 
It is a  deceptive function. 
We can prove  the monotonically increasing condition holds. 
We give an outline of the proof  but omit its detailed calculation. Corresponding to   fitness level $f_l$, the subset 
\begin{align*}
&S_{0}= \{x^* \},  \mbox{ and } f_0=n, \\
&S_l=\{x \mid h(x, x^*)=n-l \},  \mbox{ and } f_l =n-l, &\mbox{for }l>0, 
\end{align*} where $h(x,y)$ is the Hamming distance between $x$ and $y$.
  
  According to Lemma~\ref{lemExactTime}, the expected runtime of EA($\frac{1}{n}$) is given by the following recurrence relation:   $G(x)=0$ for $x \in S_0$ and 
\begin{equation*}
G(x) =
\frac{1+\sum^{l-1}_{k=0}\sum_{y \in S_k} P(x,y)G(y) }{ \sum^{l-1}_{k=0} \sum_{y \in S_k} P(x,y)}, \quad x \in S_l,
\end{equation*}
where  
\begin{align*}
P(x,y)=P^{[m]}(x,y)=\left(1-\frac{1}{n}\right)^{n-h(x,y)}  \left( \frac{1}{n}\right)^{h(x,y)}.
\end{align*}
 Then 
  the monotonically increasing condition holds.  
\begin{align*}
&G(x)< G(y) \Longrightarrow f(x)<f(y), & x,y \in S_{\non}. 
\end{align*}

  Applying Theorem~\ref{theCriteronHard}, we know the fitness function related to this instance is the hardiest in the class. 
\end{example}

\begin{example}
Consider an  instance class of the 0-1 knapsack problem. The knapsack capacity $C$ is enough large such that $C \ge w_1 + \cdots + w_n$. The unique optimum is $ (1  \cdots  1)$. This function class is equivalent to  linear functions. We apply  EA($\frac{1}{n}$) to  the problem.

Let's investigate a special instance in the class: $v_1= \cdots = v_n=1$.  
Its fitness function is equivalent to the OneMax function, so that it is easy. We  prove the OneMax function is the easiest through verifying the monotonically decreasing condition. 
We give an outline of the proof.
Corresponding to fitness level $f_l$ where $l=0, \cdots, n$, the subset  
\begin{align*}
&S_l=\{x \mid h(x, \vec{1})=l \}, &\mbox{and } f_l=n-l,
\end{align*} where $h(x,y)$ is the Hamming distance between $x$ and $y$.

  According to Lemma~\ref{lemExactTime}, the expected runtime of EA($\frac{1}{n}$) is given by the following recurrence relation:  
   $G(x)=0$ for $x \in S_0$ and for $l >0$
\begin{equation*}
G(x) =
\frac{1+\sum^{l-1}_{k=0}\sum_{y \in S_k} P(x,y)G(y) }{ \sum^{l-1}_{k=0} \sum_{y \in S_k} P(x,y)}, \quad x \in S_l,
\end{equation*}
where  
\begin{align*}
P(x,y)=P^{[m]}(x,y)=\left(1-\frac{1}{n}\right)^{n-h(x,y)}  \left( \frac{1}{n}\right)^{h(x,y)}.
\end{align*}
 Then the monotonically decreasing condition holds. 
\begin{align*}
&G(x)< G(y) \Longrightarrow f(x)>f(y), & x,y \in S. 
\end{align*}

 Applying Theorem~\ref{theCriterionEasy}, we get the OneMax function is the easiest among all linear functions.  
\end{example}

 Note: 
The  monotonically increasing condition  is   a sufficient condition for a fitness function being the hardest, but not  necessary. 
The same is true to the monotonically decreasing condition of the easiest functions. The reason is trivial: Consider a function class  only includes  one  function, then the  function will be both the easiest and hardest in the class, regardless of the monotonically increasing or decreasing condition.

\section{Construction of  Easiest and Hardest Fitness Functions to an EA}
\label{secConstruction}
In this section we   construct  unimodal  functions (the easiest) and deceptive  functions (the hardest), respectively, to any given elitist (1+1) EA.
\subsection{Construction of  Easiest  Fitness Functions}
\label{secConstructionEasy}
Given a class consisting of all   fitness functions with the same optima  on a finite set $S$, consider  an  elitist (1+1) EA   for maximising a fitness function in the class.  We construct the easiest function $f (x)$ to the EA as follows.

\begin{enumerate}
\item  Let   $S_0=S_{\opt}$.  For any $x \in S_0$, define
  $
 G'(x) =0.$

\item  Suppose that    the subsets $S_0, \cdots, S_{l-1}$ are given and $G'(x)$ has been defined on these subsets. Let $S_l$ be the set consisting of all points such that
\begin{align*}
  \arg\min_{x \in S \setminus  \cup^{l-1}_{k=0} S_k }   \frac{1+\sum^{l-1}_{k=0} \sum_{y \in S_k} P^{[m]}(x, y) G'(y)}{\sum^{l-1}_{k=0} \sum_{y \in S_k} P^{[m]}(x, y)}   .
\end{align*}
 For any $x \in S_l$, define
\begin{align}
\label{equMl}
G'(x)=    \frac{1+\sum^{l-1}_{k=0} \sum_{y \in S_k} P^{[m]}(x, y) G'(y) }{\sum^{l-1}_{k=0} \sum_{y \in S_k} P^{[m]}(x, y)}   .
\end{align}
 The value of $G'(x)$ is the same for any point $x$  in the same subset.

\item  Repeat the above step until any point is  covered by a subset. Then there exists some integer $L>0$ and 
  $ 
  S= \cup^L_{k=0} S_k.
  $ 

\item  Choose $L+1$  numbers $f_0, \cdots, f_L $ such that $f_0 >  \cdots > f_L$. Set a fitness function $f(x)$ as follows:  
$
    f(x) =f_k,  $ for $  x \in S_k.
$

\end{enumerate}

The following theorem shows that the fitness function constructed above is the easiest to the  EA. The proof is a direct application of   the monotonically decreasing condition.

\begin{theorem}\label{theEasyFunction}
$f(x)$ is the easiest function in the function class with respect to the EA.
\end{theorem}

\begin{IEEEproof} 
(1) We show that $G'(x)$   equals to the expected runtime $G(x)$.    

According to Lemma~\ref{lemExactTime},   the expected runtime  $G(x)=0$ for $x \in S_0$ and for $l >0$
\begin{align}
\label{equGForm1}
&G(x)=
\frac{1+\sum^{l-1}_{k=0}\sum_{y \in S_k} P(x,y)G(y) }{ \sum^{l-1}_{k=0} \sum_{y \in S_k} P(x,y)}, 
& x \in S_l
.
\end{align}

For any   $  x \in S_k$ and $y \in S_l$ where $k >  l$, since $f(x)=f_k  < f(y)=f_l$ and  the EA adopts  elitist selection,   $y$ is always accepted if it has been generated via mutation. Thus the transition probability $P(x,y)$ equals to $P^{[m]}(x,y)$. 
(\ref{equGForm1}) equals to
\begin{align}
\label{equGtime}
&G(x)=\frac{1+\sum^{l-1}_{k=0}\sum_{y \in S_k} P^{[m]}(x,y)G(y) }{ \sum^{l-1}_{k=0} \sum_{y \in S_k} P^{[m]}(x,y)}, & x \in S_l.
\end{align}

Comparing it with   (\ref{equMl}),  $G(x)$ and $G'(x)$ are identical.

(2) We prove   the monotonically decreasing condition.
 
First we prove  an inequality: 
\begin{align}
\label{equG1}
&G(x) > G(y), &   x \in S_{l+1}, y \in S_l.
\end{align} 

We prove it by induction.
For any $x  \in S_1$, $y \in S_0$, it is trivial that
$ 
 G(x) > G(y)=0.
$ 
Suppose that for any $x  \in S_l$, $y \in S_{l-1}$, it holds 
$ 
 G(x) > G(y) .
$ 
We   prove that for any $x \in S_{l+1}, y \in S_l$, it holds 
$ 
 G(x) > G(y) .
$ 
 
Since $y \in S_{l}$, from the construction, we know that
$$
 G(y)=   \min_{w \in S \setminus  \cup^{l-1}_{k=0} S_k }   \frac{1+\sum^{l-1}_{k=0} \sum_{z \in S_k} P^{[m]}(w, z) G(z)}{\sum^{l-1}_{k=0} \sum_{z \in S_k} P^{[m]}(w,z)}   .  
  $$
  
Let $w =x$, then  we get 
$$
G(y)\le   \frac{1+\sum^{l-1}_{k=0} \sum_{x\in S_k} P^{[m]}(x, z) G(z)}{\sum^{l-1}_{k=0} \sum_{z \in S_k} P^{[m]}(x,z)}   . 
$$
Equivalently
$$
 G(y)  \sum^{l-1}_{k=0} \sum_{z \in S_k} P^{[m]}(x,z) \le 1+\sum^{l-1}_{k=0} \sum_{z \in S_k} P^{[m]}(x,z) G(z)   .
$$

We add the term $\sum_{z \in S_l} P^{[m]}(x,z) G(z)$ to both sides. Notice that $G(z)=G(y)$ for $z  \in S_l$. As to the left-hand side, we replace the factor $G(z)$ by $G(y)$ and move it outside of the summation. Then we get
$$
G(y) \sum^{l}_{k=0} \sum_{z \in S_k} P^{[m]}(x,z) \le 1+\sum^{l}_{k=0} \sum_{z \in S_k} P^{[m]}(x,z) G(z).
$$
Equivalently
$$
 G(y)\le  \frac{1+\sum^{l}_{k=0} \sum_{z \in S_k} P^{[m]}(x,z) G(z)}{ \sum^{l}_{k=0} \sum_{z \in S_k} P^{[m]}(x,z) }.
$$

Since $x \in S_{l+1}$, it follows from (\ref{equGtime}) 
$$
 G(x)=  \frac{1+\sum^{l}_{k=0} \sum_{z \in S_k} P^{[m]}(x,z) G(z)}{ \sum^{l}_{k=0} \sum_{z \in S_k}P^{[m]}(x,z) }.
$$

So we get $ G(y) \le G(x)$.  The inequality is strict since $x$ and $y$ are in different subsets. Thus we prove (\ref{equG1}).

Secondly using (\ref{equG1}),  we can infer the monotonically decreasing condition easily. From (\ref{equG1}), we draw that 
\begin{align}
\label{equTimeInq1}
& G(x) > G(y), &\mbox{if } x \in S_{l}, y \in S_k \mbox{ with }l >k.
\end{align}

For any two points $x$ and $y$ such that $G(x) > G(y)$,  let   $x \in S_k$, $y \in S_l$. Then $k$ and $l$ must satisfy $k <l$. Then we have $
f(x)=f_k < f(y) =f_l.
$ This proves the monotonically decreasing condition.
 
(3) The conclusion is  drawn  from Theorem~\ref{theCriterionEasy}.
\end{IEEEproof}

The above theorem  provides an approach to designing the easiest fitness functions in the function class. The idea behind the construction procedure is simple: we construct a function which is unimodal in the time-based fitness landscape and then it is the easiest. Notice that the number of the easiest functions is infinite since the potential values of each $f_l$  are infinite.

\subsection{Construction of  Hardest Fitness Functions}
\label{secConstructionHard}
We  consider  an elitist (1+1) EA and a class of fitness functions  with the same optima. The hardest fitness function $f (x)$ in this class is constructed as follows.
\begin{enumerate}
\item Let   $S_0=S_{\opt}$.  For any $x \in S_0$, let
  $ 
  G'(x) =0.
  $

\item  Suppose that the subsets $S_0, \cdots, S_{l-1}$ have been produced and $G(x)$  have been defined on these subsets. Then define $S_l$ to be the set of all points   such that
  $$
  \arg\max_{x \in S \setminus  \cup^{l-1}_{k=0} S_k} \frac{1+\sum^{l-1}_{k=0} \sum_{y \in S_k} P^{[m]}(x, y) G'(y)}{\sum^{l-1}_{k=0} \sum_{y \in S_k} P^{[m]}(x, y)} .
  $$
For any $x \in S_l$, set
\begin{align}
\label{equMulb}
  G'(x)=  \frac{1+\sum^{l-1}_{k=0} \sum_{y \in S_k} P^{[m]}(x, y) G'(y)}{\sum^{l-1}_{k=0} \sum_{y \in S_k} P^{[m]}(x, y)} .
\end{align}

\item  Repeat the above step until any point is  covered by a subset. Then there exists an integer $L>0$ such that
  $ 
  S= \cup^L_{k=0} S_k.
  $ 

\item  Choose  $L+1$ number $f_0, \cdots, f_L$ such that $f_0 >  \cdots > f_L>0$. Set the fitness function   to be  
$
f(x)= f_k,   x \in S_k.
$
\end{enumerate}
 
Now we prove that $f(x)$ is the hardest fitness function in the class using the monotonically increasing condition.

\begin{theorem}\label{theHardFunction}
$f(x)$ is the hardest function in the function class to the EA.
\end{theorem}
\begin{IEEEproof} 
(1) We prove that 
the mean  runtime   $ 
G(x)=G'(x) $. The proof is similar to the first step in the proof of Theorem~\ref{theEasyFunction}.

(2) We prove the monotonically increasing condition. 
The proof is   similar to the second step in the proof of Theorem~\ref{theEasyFunction}.

(3) The conclusion is  drawn  from Theorem~\ref{theCriteronHard}.
\end{IEEEproof}

The above theorem  provides an approach to designing the hardest fitness functions in the class. We construct a function which is deceptive in the time-based fitness landscape and then it is the hardest.

In the  construction of the easiest and hardest functions,  we don't   restrict the representation of fitness functions. However, the current approach is not suitable for  the fitness function class with a specific requirement,  for example, all fitness functions in the class must be   linear or quadratic. This research issue is left for future studies.

\subsection{Case Study: Benchmarks in Pseudo-Boolean Optimisation}
\label{secCase2}
So far we have introduced a general approach to constructing the easiest and hardest fitness functions.  Now  we illustrate an application in pseudo-Boolean optimisation: to  design benchmarks within a fitness function class. According to No Free Lunch theorems, the performance of   two EAs are equivalent if averaged over all possible Boolean-valued fitness functions. Therefore we only consider a   fitness class.

\begin{example}
Consider the class of all pseudo-Boolean   functions with the same  optima at $\vec{0}:=(0  \cdots  0)$ and $\vec{1}:=(1  \cdots  1)$.   
\begin{equation}
 \max \{ f(x); x \in  \{ 0,1\}^n \}.
\end{equation}

We compare the performance of two (1+1) elitist EAs  on this problem using different mutation rates.
\begin{enumerate}
\item   \textbf{EA($\frac{1}{n}$).} Flip each bit independently with flipping probability $\frac{1}{n}$.  The mutation probability from $x$ to $y$ is
\begin{align}
P^{[m]}(x,y)=\left(1-\frac{1}{n}\right)^{n-h(x,y)}  \left( \frac{1}{n}\right)^{h(x,y)},
\end{align}
where $h(x, y)$ denote  the   Hamming distance  between   $x$ and $y$.

\item   \textbf{EA($\frac{1}{2}$).} Flip each bit independently with flipping probability $\frac{1}{2}$.  The mutation probability from $x$ to $y$ is
\begin{align}
P^{[m]}(x,y)= \left( \frac{1}{2}\right)^{n}.
\end{align}
\end{enumerate}

As to benchmark functions, their optima must be known in advance and  the number of benchmarks is   often between $5$ to $30$.  Since a function class normally includes a large amount of functions, a question is which functions should be chosen as benchmarks? Naturally we prefer  typical functions in the class: easy, hard and  `averagely hard' functions. Here we only consider how to design the easiest and hardest fitness functions.   

The easiest fitness function  to EA($\frac{1}{n}$) is constructed  as follows.

\begin{enumerate}
\item  Let   $S_0=\{\vec{0}, \vec{1} \}$.  For any $x \in S_0$, define
  $
 G(x) =0.$

\item  Suppose that    the subsets $S_0, \cdots, S_{l-1}$ are given and $G(x)$ has been defined on these subsets. Let $S_l$ be the set consisting of all points such that
\begin{align}
\label{equA1}
  \arg\min_{x \in S \setminus  \cup^{l-1}_{k=0} S_k }   \frac{1+\sum^{l-1}_{k=0} \sum_{y \in S_k} P^{[m]}(x, y) G(y)}{\sum^{l-1}_{k=0} \sum_{y \in S_k} P^{[m]}(x, y)}   .
\end{align}
  Using the mutation probability
  \begin{align*}
P^{[m]}(x,y)=\left(1-\frac{1}{n} \right)^{n-h(x,y)}  \left( \frac{1}{n}\right)^{h(x,y)},
\end{align*}
we get
\begin{align*}
S_l=\{x \mid \min\{ h(x, \vec{0}), h(x,\vec{1})\}=l \}.
\end{align*}

 For any $x \in S_l$, define
\begin{align}
\label{equRecurrenceEasy}
G(x)=    \frac{1+\sum^{l-1}_{k=0} \sum_{y \in S_k} P^{[m]}(x, y) G(y) }{\sum^{l-1}_{k=0} \sum_{y \in S_k} P^{[m]}(x, y)}   .
\end{align}

  \item Repeat the above step until any point is  covered by a subset. The last subset is $S_L$ where $L:=  n/2$. Without loss of generality, assume $n$ is even.
  
 \item   Choose $L+1$  numbers
$f_0,   \cdots, f_L$ such that $f_0 > \cdots >f_L$. Set the fitness function 
$
 f(x)= f_l,  $ for $x \in S_l.
$ Then $f(x)$ is the easiest function  in the function class.

\end{enumerate}

An example  of the easiest function to EA($\frac{1}{n}$) is the Two Max function, given by
\begin{align}
\label{equTwoMax}
f(x)= n-\min\{ h(x, \vec{0}), h(x,\vec{1})\}.
\end{align} 
The runtime is calculated as follows. Let $x \in S_l$, without loss of generality, suppose it has $l$ 0-valued bits and $n-l$ 1-valued bits (with $l \le n-l$). The event of going from the fitness level $f_l$ to a higher fitness level will happen if one of 0-valued bits is flipped and other bits are kept unchanged. The probability of this event is at least
$$
\binom{l}{1}\frac{1}{n}\left(1-\frac{1}{n}\right)^{n-1} \ge \frac{l}{ne},
$$where $e$ is Euler's constant. 
Thus the runtime  of going from the fitness level $f_l$ to a higher fitness level is no more than
$
\frac{en}{l}.
$
 Since the number of fitness levels is   $L$, therefore the total runtime to reach the global optima
 is at most
 \begin{align*}
 \sum^L_{l=1} \frac{en}{l} =O(n \ln n).
 \end{align*} 

There are infinite easiest fitness functions, including linear, quadratic and other  non-linear functions, for example,
 \begin{align}
&f(x)= \left(n-\min\{ h(x, \vec{0}), h(x,\vec{1})\}\right)^k, & k=1, 2, \cdots
\end{align} 

The runtime of the EA on all easiest fitness functions is the same no matter whether they are linear or not.

It is worth noting that the Two Max function is unimodal   in the time-based fitness landscape. But using the Hamming distance, the function is two-modal due to two optima at $\vec{0}$ and $\vec{1}$.   

The hardest fitness function to EA($\frac{1}{n}$)  is constructed as follows. 
\begin{enumerate}
\item Let   $S_0=\{\vec{0}, \vec{1} \}$.  For any $x \in S_0$, let
  $ 
  G(x) =0.
  $

\item  Suppose that the subsets $S_0, \cdots, S_{l-1}$ have been produced and $G(x)$  have been defined on these subsets. Then define $S_l$ to be the set of all points   such that
  $$
  \arg\max_{x \in S \setminus  \cup^{l-1}_{k=0} S_k} \frac{1+\sum^{l-1}_{k=0} \sum_{y \in S_k} P^{[m]}(x, y) G(y)}{\sum^{l-1}_{k=0} \sum_{y \in S_k} P^{[m]}(x, y)} .
  $$
  Using the mutation probability
  \begin{align*}
P^{[m]}(x,y)=\left(1-\frac{1}{n}\right)^{n-h(x,y)}  \left( \frac{1}{n}\right)^{h(x,y)},
\end{align*}
 we get (let $L:= n/2$ and assume $n/2$ is an integer)
\begin{align*}
S_{l}=\{x \mid \min\{ h(x, \vec{0}), h(x,\vec{1})\}=L-l-1 \}.
\end{align*}

For any $x \in S_l$, set
\begin{align}
\label{equRecurrenceHard}
  G(x)=  \frac{1+\sum^{l-1}_{k=0} \sum_{y \in S_k} P^{[m]}(x, y) G(y)}{\sum^{l-1}_{k=0} \sum_{y \in S_k} P^{[m]}(x, y)} .
\end{align}

 \item Repeat the above step until any point is  covered by a subset. The last subset is $S_L$.
  
 \item   Choose $L+1$  numbers
$f_0,   \cdots, f_L$ such that $f_0 > \cdots >f_L$. Set the fitness function 
$
 f(x)= f_l,  $ for $x \in S_l.
$
 \end{enumerate}

An example of the hardest function to EA($\frac{1}{n}$) is  a Fully Deceptive function
\begin{equation}
\label{equFullyDec}
f(x)=\left \{
\begin{array}{llll}
  n+1, & \mbox{if }  x = \vec{0},\vec{1};\\
\min\{h(x,\vec{0}),h(x, \vec{1})\}, &\mbox{otherwise}.  
\end{array}
\right.
\end{equation}
 
Consider a point $x \in S_1$ where $x$ consists of exact $n/2$ zero-valued bits and $x$ is the farthest  from   $\vec{0}$ and $\vec{1}$. Now we calculate the runtime $G(x)$. Since the Hamming distance between $x$ and the optima $\vec{0}$ and $\vec{1}$ is  $n/2$, so the transition probability of going from $x$ to the two optima is between
\begin{align*}
\left(1-\frac{1}{n}\right)^{n/2}  \left( \frac{1}{n}\right)^{n/2} 
 \mbox{ and }2\left(1-\frac{1}{n}\right)^{n/2}  \left( \frac{1}{n}\right)^{n/2},
\end{align*}
and the runtime is
$
 \Theta(n^{n/2}).
$

There are infinite hardest fitness functions,   for example, for $k=1, 2, \cdots$
 \begin{align}
f(x)=\left \{
\begin{array}{llll}
  (n+1)^k, & \mbox{if }  x = \vec{0},\vec{1};\\
\left(\min\{h(x,\vec{0}),h(x, \vec{1})\}\right)^k, &\mbox{otherwise}.  
\end{array}
\right. 
\end{align}

We can construct the easiest and hardest fitness functions to EA($\frac{1}{2})$ in the same way.
The easiest fitness function  to EA($\frac{1}{2}$) is constructed  as follows.
\begin{enumerate}
\item Let $S_0$ be the set of optima $\vec{0}$ and $\vec{1}$.  

\item  
Let $S_1$  be the set consisting of all points such that
  $$
  \arg\min_{x \in S \setminus S_0}   \frac{1}{\sum_{y \in S_0} P^{[m]}(x, y)}.
  $$
Using the mutation probability
  \begin{align*}
P^{[m]}(x,y)= 2^{-n},
\end{align*}
 we get $
S_1=\{x \mid x \neq \vec{0}, \vec{1}\}.
$

 \item   Choose $2$  numbers
$f_0,     f_1$ such that $f_0  >f_1$. Set the fitness function 
$
 f(x)= f_l,  $ for $x \in S_l.
$ Then $f(x)$ is the easiest function  in the function class.
\end{enumerate}

An example of the easiest function to EA($\frac{1}{2}$) is the Two Needles in the Haystack function
\begin{equation}
\label{equNeedles}
f(x)=\left \{
\begin{array}{llll}
1, & \mbox{if }  x = \vec{0},\vec{1};\\
0, &\mbox{otherwise}.  
\end{array}
\right.
\end{equation}

We calculate the runtime $G(x)$ for $x \in S_1$ as follows. The transition probability of going from $x$ to the two optima is between $
  (\frac{1}{2})^{n} 
$ and 
$
2 \times (\frac{1}{2})^{n}.
$
Then the runtime is $\Theta(2^n)$.
 
The hardest fitness function to EA($\frac{1}{2}$)  is constructed as follows. 
\begin{enumerate}
\item Let   $S_0=\{\vec{0}, \vec{1}\}$.  

\item  Let $S_1$   be the set of all points such that
  $$
  \arg\max_{x \in S \setminus S_0 }   \frac{1}{\sum_{y \in S_0} P^{[m]}(x, y)}.
  $$
  
   Using the mutation probability
  \begin{align*}
P^{[m]}(x,y)= 2^{-n},
\end{align*}
 we get $
S_1=\{x \mid x \neq \vec{0}, \vec{1} \}.
$

 \item   Choose 2 numbers
$f_0,   f_1$ such that $f_0  >f_1$. Set the fitness function 
$
 f(x)= f_l,  $ for $x \in S_l.
$ 
Then the above function $f(x)$ is the  hardest to EA($\frac{1}{2}$). 
 \end{enumerate}

An example of the hardest function to EA($\frac{1}{2}$) is the   Two Needles in the Haystack function, the same as the easiest function. 
The runtime is $\Theta(2^n)$.  Since the runtime of EA($\frac{1}{2}$) on both the easiest and hardest functions is $\Theta(2^n)$. Then we know for any function in the class, its runtime is $\Theta(2^n)$.
 
We have constructed three benchmark functions:  Two Max, Fully Deceptive and Two Needles in the Haystack. They are described in Table~\ref{tab1}. The three functions represent three typical fitness landscapes: unimodal, deceptive and isolation. Using the   benchmarks, we can make a fair comparison of  the  performance of EA($\frac{1}{n}$) and EA($\frac{1}{2}$).  
Table~\ref{tab1} lists the results.

\begin{table*}[!t]
\caption{Three benchmarks and Runtime Comparison of Two EAs.}
\label{tab1}
\centering
\begin{tabular}{ccccc}
\hline
name &function & time-based fitness landscape &EA($\frac{1}{n}$)  & EA($\frac{1}{2}$)\\
\hline
 Two Max & $f(x)=n-\min\{ h(x, \vec{0}), h(x,\vec{1})\}$ &  unimodal &$O(n \ln n)$ & $\Theta(2^n)$\\
 \hline
Fully Deceptive  & $f(x)=\left \{
\begin{array}{llll}
  n+1, & \mbox{if }  x = \vec{0},\vec{1};\\
\min\{h(x,\vec{0}),h(x, \vec{1})\}, &\mbox{otherwise}.  
\end{array}
\right.$ & deceptive & $\Theta(n^{n/2})$  & $\Theta(2^n)$\\
\hline
 Two Needles in Haystack &  $f(x)=\left \{
\begin{array}{llll}
1, & \mbox{if }  x = \vec{0},\vec{1};\\
0, &\mbox{otherwise}.  
\end{array}
\right.$ & isolation &  $n^{\Theta(n)} \times \Theta(1)$ &  $\Theta(2^n)$\\
\hline
\end{tabular}
\end{table*}

The runtime of EA($\frac{1}{n}$) on the Two Needles in the Haystack function is calculated as follows. Suppose the initial point $x$ consists of $\Theta(n)$ 0-valued bits and $\Theta(n)$ 1-valued bits, then the event of going from $x$ to the optima happens   when either all $0$-valued bits are flipped and other bits unchanged; or all $1$-valued bits are flipped and other bits unchanged. The probability of the event is   $  (\frac{1}{n})^{\Theta(n)} \times \Theta(1)$. Thus the runtime is $n^{\Theta(n)}\times \Theta(1)$.

 From the table, we see  that  EA($\frac{1}{n}$) is better   than   EA($\frac{1}{2}$) on the Two Max function, but worse on the Fully Deceptive Points and Two Needles in the Haystack functions. The comparison gives an understanding of the two EAs' ability in  different fitness landscapes: unimodal, deceptive and isolation. Each EA has its own advantage.   EA($\frac{1}{n}$) is more suitable for unimodal functions, but EA($\frac{1}{2}$) performs better on deceptive or isolation functions.

The runtime of EA($\frac{1}{n}$) and EA($\frac{1}{2}$) increases exponentially fast on the Fully Deceptive and Two Needles in the Haystack functions. Thus it will be difficult to compare the runtime of the EAs  via computer experiments unless $n$ is   small. 
\end{example}

\section{Mutual Transformation Between the Easiest and Hardest Fitness Functions}
\label{secMutual}
In the   case study of the previous section, we  observe that the easiest and hardest fitness functions vary as EAs change. In this section we prove an interesting result: a fitness function that is the easiest to one elitist (1+1) EA could becomes  hardest to another  elitist (1+1) EA and vice versa.   

\subsection{Easiest May Become Hardest}
\label{secEasyToHard}
Consider a class consisting of all   functions with the same optima. Let $f(x)$ be the easiest to an elitist (1+1) EA (called the \emph{original EA}). Denote its fitness levels by $f_0 >   \cdots >f_L $  and define the set $S_l=\{ x; f(x)=f_l\}$.  
We construct another  elitist  (1+1) EA  (called the \emph{bad EA}) and show $f(x)$ is  the hardest to  the bad EA.   

The   mutation operator in the bad  EA is constructed as follows.
\begin{enumerate} 
\item Choose $L+1$ non-negative numbers $ m_0, m_1,  \cdots, m_L  $ such that
$
m_0=0, m_1 >m_2 >\cdots > m_L>0.
$

\item For  any  $x \in S_0$ and $y \in S$, let the mutation transition probability  
 $P^{[m]}(x, y) $ be any probability.

\item 
For any $x \in S_{l}$ (where $l =1, \cdots, L$) and $y \in S$, set the mutation transition probability $P^{[m]}(x, y) $ such that
 \begin{align*} 
    &  \frac{1+\sum^{k-1}_{j=0} \sum_{y \in S_j} P^{[m]}(x, y) m_j}{\sum^{k-1}_{j=0} \sum_{y \in S_j} P^{[m]}(x, y)}  < m_{k}, &\mbox{for } k <l, 
    \end{align*}
    and
    \begin{align}
    &\frac{1+\sum^{l-1}_{j=0} \sum_{y \in S_j} P^{[m]}(x, y) m_j}{\sum^{l-1}_{j=0} \sum_{y \in S_j} P^{[m]}(x, y)}  = m_{l}. \label{equMutPro1}
\end{align}
\end{enumerate} 
 The above mutation operator is determined by the subsets $S_1, \cdots, S_l$ rather than fitness levels.

The following theorem shows the function $f(x)$ satisfies the monotonically increasing condition and then it is the hardest  to the bad EA. 

 \begin{theorem}\label{theEasy2Hard}
   $f(x)$ is the hardest function to the bad EA.
\end{theorem}
\begin{IEEEproof}
(1) We prove that
the expected runtime of  the bad EA  
$
G(x)=m_l, $ for $ x \in S_l, l=0,  \cdots, L.
$
 
  According to Lemma~\ref{lemExactTime},   the expected runtime  
\begin{align*}
 G(x)= \frac{1+\sum^{l-1}_{j=0} \sum_{y \in S_j} P  (x, y) G(y)}{\sum^{l-1}_{j=0} \sum_{y \in S_j} P (x, y)}.
\end{align*}

For any   $  x \in S_l$ and $y \in S_k$ where $l >k $. Since $f(x)=f_l < f(y)=f_k$ and  the bad EA adopts  elitist selection,   $y$ is always accepted if it has been generated via mutation. Thus the transition probability $P(x,y)$ equals to $P^{[m]}(x,y)$. 

The expected runtime  becomes
\begin{align*}
 G(x)= \frac{1+\sum^{l-1}_{j=0} \sum_{y \in S_j} P^{[m]}(x, y) G(y)}{\sum^{l-1}_{j=0} \sum_{y \in S_j} P^{[m]}(x, y)}.
\end{align*}

Comparing it with   (\ref{equMutPro1}), we obtain $G(x)$ and $m_l$ are identical.
 
(2) We prove the monotonically increasing condition.
 
Assume that   $x \in S_l$, $y \in S_k$  for some $l$ and $k$. If $  G(x) < G(y)$, then it is equivalent to $m_l < m_k$. Thus we have $k  <l$ and 
\begin{align*}
f(x)=f_l < f(x)=f_k.
\end{align*}
which gives the  monotonically increasing condition.

(3)  The conclusion is drawn  from Theorem~\ref{theCriteronHard}.
\end{IEEEproof}

 In the construction of the mutation operator and the proof of the above theorem, we  don't utilize the assumption of $f(x)$ being the easiest to the original EA. Thus the theorem  can be understood more generally:  for any  fitness function $f(x)$, we can construct an elitist (1+1) EA to which $f(x)$ is  the hardest. From the theoretical viewpoint, the theorem shows the existence of a bad EA to the easiest   fitness function.

\subsection{Hardest May Become Easiest}
\label{secHardToEasy}
Let $f(x)$ be   the hardest fitness function to the original elitist (1+1) EA. We construct another  elitist  (1+1) EA (called the \emph{good EA}), and show $f(x)$ becomes the easiest to the good EA.  

The   mutation operator in the good EA  is constructed as follows.
\begin{enumerate} 
\item Choose $L+1$ non-negative numbers $m_0,   \cdots, m_L$ such that
$
m_0=0<   \cdots< m_L.
$

\item For  any  $x \in S_0$ and $y \in S$, let the mutation transition probability  
 $P^{[m]}(x, y) $ be any probability.

\item 
For any $x \in S_{l}$ (where $l =1, \cdots, L$) and $y \in S$, set the mutation transition probability $P^{[m]}(x, y) $ such that
 \begin{align*} 
    &  \frac{1+\sum^{k-1}_{j=0} \sum_{y \in S_j} P^{[m]}(x, y) m_j}{\sum^{k-1}_{j=0} \sum_{y \in S_j} P^{[m]}(x, y)}  > m_{k}, &\mbox{for } k <l, 
    \end{align*}
    and
    \begin{align}
    &\frac{1+\sum^{l-1}_{j=0} \sum_{y \in S_j} P^{[m]}(x, y) m_j}{\sum^{l-1}_{j=0} \sum_{y \in S_j} P^{[m]}(x, y)}  = m_{l}. \label{equMutPro4}
\end{align}

\end{enumerate} 

The following theorem shows $f(x)$  satisfies monotonically decreasing condition and then it is the easiest  to the good EA.
  
\begin{theorem}\label{theHard2Easy}
   $f(x)$ is the easiest function to the good EA.
\end{theorem}

\begin{IEEEproof} 
(1) We prove that the expected runtime of   the good EA   $
G(x)=m_k, $ for $ x \in S_k, k=0,1, \cdots, L.$
 The proof is similar to the first step in the proof of Theorem~\ref{theEasy2Hard}. 

(2) We prove  the monotonically decreasing condition.
 The proof is similar to the second step in the proof of Theorem~\ref{theEasy2Hard}.

(3) The conclusion is drawn   from Theorem~\ref{theCriterionEasy}.
\end{IEEEproof}

 In the construction of the mutation operator and the proof of the above theorem, we also don't utilize the assumption of $f(x)$ being the hardest to the original EA. The theorem  implies that  for any  fitness function $f(x)$, we can construct a good (1+1) EA to which $f(x)$ is  the easiest.

The above theorem reveals  if a fitness function is the hardest to one EA, then it is possible to design another good  EA to which the function is the easiest.  However,  the above construction method is intractable in practice since the complexity of construction is exponential. How to design such a good EA  is an ultimate goal in the study of EAs but beyond the scope of the current paper.

  Theorems~\ref{theEasy2Hard} and~\ref{theHard2Easy} can be viewed as   a complement to No Free Lunch theorems. No Free Lunch theorems concern all potential fitness functions. The theorems claim the performance of any two EAs are equivalent if averaged over all possible functions. Theorems~\ref{theEasy2Hard} and~\ref{theHard2Easy} concern the hardness of a single fitness function. The two theorems  assert that a fitness function could be the easiest to one elitist (1+1) EA but  the hardest to another EA.  This implies for a single fitness function, a good EA (but also a bad EA)  always exists.    

\section{Conclusions and Future Work}
\label{sec9}
This paper presents a rigorous analysis devoted to the easiest and hardest fitness functions  with respect to  any  given elitist  (1+1) EA  for maximising a class of fitness functions with the same optima. Such fitness functions have been constructed step by step. It is demonstrated that the unimodal functions are  the easiest and deceptive functions are the hardest in terms of the time-based fitness landscape.
Furthermore it  reveals that the hardest (and easiest) functions may become the easiest (and hardest) with respect to another elitist (1+1) EA.   From the theoretical viewpoint,  a good EA (but also a bad EA)  always exists  for a single fitness function.

A potential application of the theoretical work is the design of benchmarks. Benchmarks play  an essential role in the empirical comparison of  EAs. In order to make a fair comparison,  a good practice is to choose typical fitness functions in benchmarks, for example, several easy, hard and `averagely hard' fitness functions. Our work provides a theoretical guideline to the design of easy and hard functions: to choose    unimodal (the easiest) and  deceptive (the hardest) fitness functions with respect to   EAs under comparison. 

Another   application   is to understand the ability of EAs on a class of fitness functions with the same optima.  Through the comparison of EAs on the easiest and hardest fitness functions, our work helps  understand the ability of  EAs in unimodal and deceptive time-based fitness landscapes.  
This has been shown in the second case study.

Non-elitist EAs, population-based EAs and dynamical EAs are not investigated in this paper. The extension of our work to such EAs will be the future research.   Another work in the future  is to study how to construct the easiest and hardest fitness functions such that a special requirement, for example, all fitness functions must be linear or quadratic.

\end{document}